%% file: main.tex
\documentclass{opt2026}

\usepackage{booktabs}
\usepackage{microtype}

\newcommand{\Zp}{\mathbb{Z}_p}
\newcommand{\gap}{\delta^2}

\title[Structured Features Overfit Where Random Features Grok]{Structured Features Overfit Where Random Features Grok}

\optauthor{%
\Name{Chon-Fai Kam} \Email{dubussygauss@gmail.com}\\
\addr University Paris City \& University of Reunion, Paris, France;
Dipartimento di Fisica e Chimica Emilio Segr\`e, Universit\`a degli Studi di Palermo, Italy
\AND
\Name{Miloud Bessafi} \Email{miloud.bessafi@gmail.com}\\
\addr EnergyLab, University of Reunion, Saint-Denis, France
\AND
\Name{Frederic Cadet} \Email{frederic.cadet.run@gmail.com}\\
\addr University Paris City \& University of Reunion, Paris, France;
PEACCEL, AI for Biologics, Paris, France%
}

\begin{document}

\maketitle

\input{body}

\end{document}

%% file: body.tex
\begin{abstract}
Xu, Vardi and Safran (ICML 2026) prove that over-parameterized ridge regression over an unstructured random Gaussian feature map groks, with the delay between memorization and generalization growing as $1/\lambda$ in the weight decay. We show that on a structured feature map the same delay does not appear. For a band-limited Fourier feature map over $\Zp^2$ carrying a single-character target that lies inside the expressible class, enlarging the band at fixed positive weight decay drives peak held-out accuracy monotonically from $1.00$ to $0.07$, with no memorize-then-generalize regime anywhere along the sweep. The degradation is not an interpolation effect. It sets in at capacity ratio $q/n = 0.638$, far below the interpolation threshold, on separate grounds from the exact null space that appears above it. What does have a sharp boundary is the active support. Holding the nominal dimension fixed and masking the band back to $1089$ active modes restores held-out accuracy of $1.000$ with zero variance across seeds, while the full $4225$-mode band collapses to $0.185$. The number of active modes acts through the teacher-weighted spectrum of the empirical Gram matrix and not through the capacity ratio, which makes this a statement about feature geometry and not a restatement of double descent.
\end{abstract}

\section{Introduction}

Grokking, the onset of generalization long after a network has fit its training data, was first observed by \citet{power2022} on small algorithmic datasets. On modular arithmetic it has since been traced to the emergence of a Fourier-based algorithm in the trained weights~\citep{nanda2023}, to a representation-learning transition~\citep{liu2022}, and to a shift from lazy to rich training dynamics~\citep{kumar2024}. In at least one clean setting it is now understood as a consequence of over-parameterization. \citet{xvs2026} (XVS) prove an end-to-end grokking guarantee for over-parameterized ridge regression trained by gradient descent with weight decay, and give quantitative bounds on the generalization delay $t_2 - t_1$ in terms of the training hyperparameters, with $t_2 \propto 1/\lambda$ as the weight decay $\lambda \to 0$. Their analysis holds for any realizable teacher over \emph{any fixed feature map}, and their experiments use an unstructured feature map, either the identity or i.i.d.\ Gaussian random features. In that regime the null space created by over-parameterization is isotropic. Every extra dimension carries a little signal, and weight decay resolves the delayed transition that defines grokking.

A natural question is whether the same picture holds when the feature map is \emph{structured} and not random. Kam et al.'s solvable holomorphic model~\citep{kam_holo} shows that a modular target is representable if and only if its discrete Fourier support lies inside the class the architecture can express. We work with a symmetric-Laurent-band extension of that class (Section~\ref{sec:setup}), which enlarges it from a positive Fourier line to a signed band so that rules such as $a-b$ become representable. That raises the dynamical sequel. Once a target lies \emph{inside} the band, does training find it immediately, only after memorization, or not at all?

The natural expectation is grokking. XVS establish that a fixed feature map produces the memorize-then-generalize delay whenever the problem is over-parameterized and realizable, and a band-representable target on an over-parameterized band feature map appears to satisfy both conditions, so it should be memorized first and generalized later. They do not grok. At fixed positive weight decay, enlarging the band produces overfitting collapse and no delayed transition, and the outcome follows from which modes are active, not from the capacity ratio. We characterize an absence, meaning the conditions under which the memorize-then-generalize delay fails to appear, and trace it to a structural property of the feature map.

Representability on this feature map is decidable in closed form, inherited from the exact Fourier analysis of Kam's holomorphic model and verified numerically at $\gap \sim 10^{-12}$ for representable targets and $\gap = 1$ for the full-plane target $ab$ (Section~\ref{sec:setup} and Appendix~\ref{app:closedform}), so the static question is settled before any dynamics are run. Enlarging the band for a representable target then degrades generalization monotonically instead of inducing grokking, with held-out accuracy falling from $1.00$ to near chance as $q/n$ grows. A masking experiment at fixed nominal dimension isolates the active support as the variable that carries the effect (Section~\ref{sec:overfit}).

The band-limited character map is the right object for this test, not just a convenient one. XVS assume realizability over a fixed feature map, and on an unstructured map that assumption cannot be audited. Here the expressible class is a span of group characters, so realizability is decided by a Fourier support condition in closed form, with no training run. We can hold the hypothesis of their theorem fixed and vary only the geometry of the features, which is the comparison the question demands.

The structure of a feature map is a control variable for whether grokking occurs at all, and not a detail of the experimental setting.

\section{Setup}
\label{sec:setup}

\paragraph{Feature map.} Fix an odd prime $p$ and $\omega = e^{2\pi i / p}$. A modular pair $(a,b) \in \Zp^2$ is encoded through its Laurent harmonics: for a band radius $M$, the feature vector collects the characters $\chi_{r,s}(a,b) = \omega^{ra+sb}$ for $-M \le r,s \le M$, realized in real coordinates as $\cos$ and $\sin$ channels. The band is symmetric, $B_M = -B_M$, and $p$ is odd, so the conjugate pair $\pm(r,s)$ satisfies $\cos\langle -r{,}-s,\theta\rangle = \cos\langle r{,}s,\theta\rangle$ and $\sin\langle -r{,}-s,\theta\rangle = -\sin\langle r{,}s,\theta\rangle$. Each pair contributes two independent real functions, not four, and the $2(2M+1)^2$ implemented channels span a function space of dimension only
\begin{equation}
q = (2M+1)^2 .
\label{eq:dim}
\end{equation}
The design matrix has numerical rank $q$ at every $M$ tested. We report the capacity ratio as $q/n$ throughout, since it is $q$, and not the number of implemented columns, that decides interpolation. Every feature entry has unit modulus. The symmetric Laurent band is introduced here. It admits negative frequencies through the two-channel realification $\chi = A + iB$, $\overline{\chi} = A - iB$ with $A=\cos$ and $B=\sin$, and closes to $kB_M = [-kM,kM]^d$ under a degree-$k$ monomial. This real, signed-frequency realization is what places the model inside the real-valued ridge framework we use, and what makes signed rules such as $a-b$ representable where a positive Fourier line would exclude them.

\paragraph{Representability.} A target $f$ is \emph{representable} iff its Fourier support lies inside the band $A = \{(r,s):|r|,|s|\le M\}$, i.e.\ iff the spectral gap
\begin{equation}
\gap := \sum_{(u,v)\notin A} |\widehat{\omega^f}(u,v)|^2 = 0.
\end{equation}
Because the features are group characters, $\gap$ is a static, training-free quantity, decidable in closed form with no training run, a property the Gaussian feature map does not have. A linear rule $ma+nb$ is a single Fourier mode, representable as soon as $|m|,|n|\le M$, so the signed rule $a-b$ = mode $(1,-1)$ is representable at $M\ge 1$, confirmed numerically at $\gap \sim 10^{-12}$. The multiplication rule $ab$ has full-plane Fourier support and is representable at no finite $M$ (a consequence of the quadratic Gauss sum). Every failure reported below is a failure to \emph{find} a rule the class contains, never a failure to contain it. The derivation and the numerical check across targets and band radii are in Appendix~\ref{app:closedform}. The closed-form structure of modular-arithmetic solutions has been studied for two-layer networks by \citet{gromov2023} and extended to modular polynomials by \citet{doshi2024}. Our band feature map inherits this Fourier structure but freezes it into a fixed, checkable expressible class~\citep{kam_holo}.

\paragraph{Model and training.} We train a linear read-out $\hat y = \Phi\theta$ over the fixed feature map, minimizing mean-squared error with $\ell_2$ regularization (ridge), by gradient descent with weight decay $\lambda$. The target phase $\omega^{f(a,b)}$ is regressed in $(\cos,\sin)$ coordinates and decoded by angle. We train on a fraction of the $p^2$ pairs and evaluate held-out accuracy. We use $p=97$ throughout. We train only the linear read-out, because the monomial activation of the parent holomorphic model acts as a frequency dilation ($B_M \to kB_M$) and is not a trainable object. The two-layer non-convex case is outside our scope (Section~\ref{sec:scope}). Following XVS, $t_1$ is the step at which training accuracy first exceeds threshold and $t_2$ the step at which held-out accuracy does. A positive gap $t_2 - t_1$ is the grokking signature. Full experimental detail, seeds, and code availability are in Appendix~\ref{app:repro}.

\section{Over-parameterization overfits, it does not grok}
\label{sec:overfit}

We train the linear read-out by gradient descent and vary the band radius $M$, equivalently the capacity ratio $q/n$ of Eq.~\eqref{eq:dim}, for the representable target $a-b$, holding $\lambda = 10^{-2}$ fixed. With $n = 3763$, over three seeds (Figure~\ref{fig:main} left, values in Appendix~\ref{app:repro}).

\begin{figure}[h]
\centering
\includegraphics[width=0.9\textwidth]{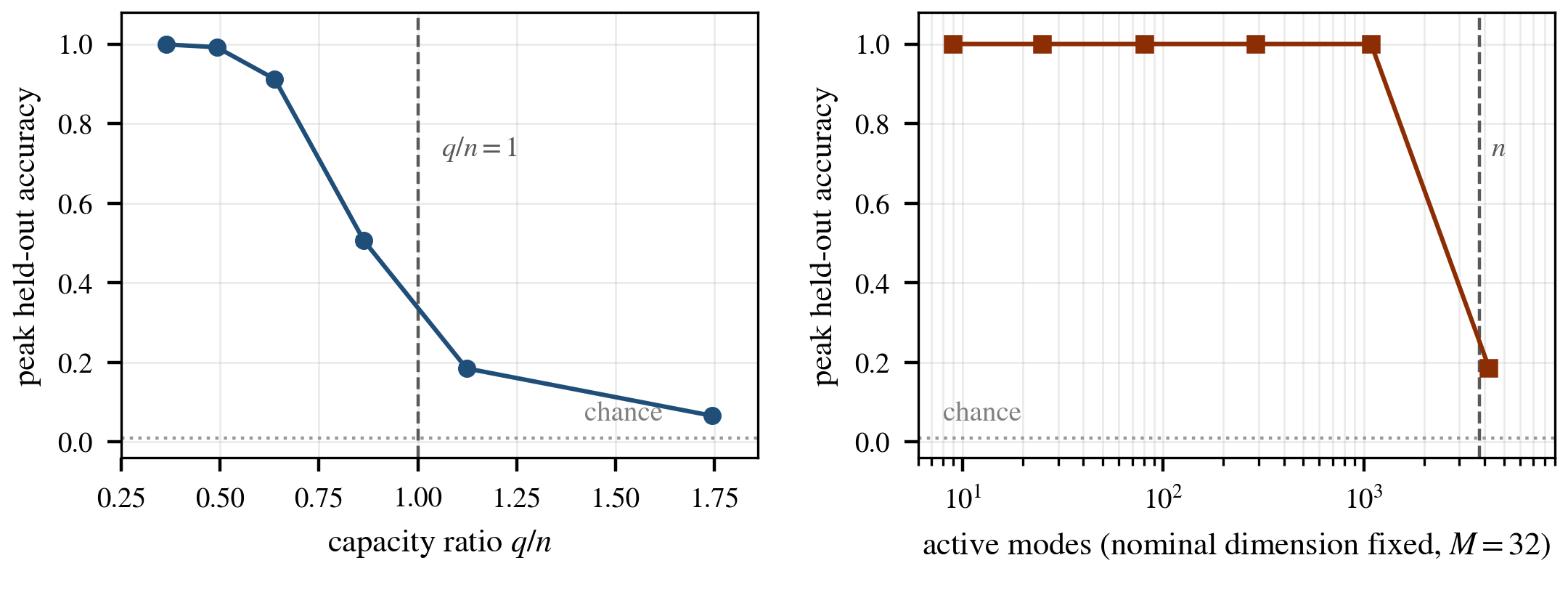}
\caption{\textbf{Left.} Peak held-out accuracy versus the capacity ratio $q/n$ for the representable target $a-b$, with the interpolation threshold and the chance level $1/97$ marked. Accuracy has already fallen to $0.506$ at $q/n=0.863$, to the left of the threshold. \textbf{Right.} Peak held-out accuracy versus the number of active modes at fixed nominal dimension, $M=32$, with $n$ marked. Restricting the active support to $1089$ modes gives perfect and zero-variance generalization, while the full band collapses. Both panels use the functional dimension of Eq.~\eqref{eq:dim} and plot the values of Tables~\ref{tab:collapse} and~\ref{tab:mask}. Three seeds, with error bars where visible.}
\label{fig:main}
\end{figure}

The XVS picture predicts the opposite. There, deeper over-parameterization amplifies grokking, so the delay grows and generalization eventually succeeds. Here it \emph{destroys} generalization, and peak held-out accuracy falls monotonically to $0.066$ at $q/n \approx 1.7$, barely above the chance level $1/97$.

Generalization is badly degraded at $M=24$ and $M=28$, where $q/n = 0.638$ and $0.863$ and the model is \emph{under}-parameterized. The degradation begins below the interpolation threshold and cannot be attributed to interpolation. Only from $M=32$ does $q$ exceed $n$, forcing a null space of dimension $q-n$. The table is consistent with two regimes and not one transition, attenuation of small positive directions below the threshold and an irreducible floor above it, which we do not attempt to separate. The $M=24$ row resembles grokking, with training accuracy saturating at $1.00$ while held-out accuracy stalls at $0.91$. It is not a delay, and the reason is structural. At $M=24$ we have $q < n$, so $\Phi_{\mathrm{tr}}$ has full column rank $q$ and the orthogonal complement of its row space inside the $2q$ implemented coordinates is precisely the gauge redundancy of the duplicated realification of Eq.~\eqref{eq:dim}. Those directions carry no function at all, so the correct solution that weight decay is supposed to surface cannot be hiding in them. Appendix~\ref{app:transient} gives the rank audit and the trajectory.

\paragraph{The mechanism: which modes are active.} Why does the structured feature map overfit where the Gaussian one groks? The extra dimensions are not isotropic. The target $a-b$ occupies one Fourier mode, every other band mode is orthogonal to it on the population, and only the finite sample mixes them into the target direction. We test the role of the active support by fixing the nominal dimension at $M=32$ and masking all but the modes within radius $k$ of the origin (Figure~\ref{fig:main} right).

At $k=16$ the model is nominally $4225$-dimensional but effectively $1089$-dimensional and generalizes perfectly, while at $k=32$ the same nominal dimension collapses. Over-parameterization \emph{per se} is not the cause. A large active band \emph{of signal-free structured modes} is.

The count is a capacity proxy rather than the exact variable. Take an in-band single-character teacher $e_t$ and let $G = \Phi_{\mathrm{tr}}^\dagger\Phi_{\mathrm{tr}}/n$ be the empirical character Gram matrix. The terminal population error of the ridge fit is then the teacher-weighted spectral integral
\begin{equation}
\lambda^2\, e_t^\dagger (G+\lambda I)^{-2} e_t \;=\; \int \Big(\frac{\lambda}{\mu+\lambda}\Big)^{\!2} \mathrm{d}\nu^G_t(\mu),
\label{eq:ridgelaw}
\end{equation}
where $\nu^G_t$ is the spectral measure of $G$ viewed from the teacher (immediate from the normal equations and diagonalization). The number of active modes enters only through $\nu^G_t$, since enlarging the active support moves teacher-weighted mass towards zero, where the factor $(\lambda/(\mu+\lambda))^2$ attenuates it. Masking changes that measure decisively. It does not make the count the control variable, and a formula in $(q,n,\lambda)$ alone cannot be exact for every teacher.

Equation~\eqref{eq:ridgelaw} also settles something the accuracy curves alone cannot, namely whether the collapse is an artefact of a finite training budget. It is not. The quantity in Eq.~\eqref{eq:ridgelaw} is the error at the \emph{fixed point} of the regularized objective and not an error after some number of steps, so at positive $\lambda$ it bounds what any amount of further training can reach. Once teacher-weighted mass sits at eigenvalues $\mu \ll \lambda$, the factor $(\lambda/(\mu+\lambda))^2$ leaves that mass almost untouched and the optimum itself is wrong. Gradient descent converging to that fixed point converges to a predictor that never generalizes, which is why the delay is absent and not long.

\section{Discussion}
\label{sec:scope}

In the XVS setting the feature map is random and isotropic, so the null-space directions created by over-parameterization are statistically aligned with the target, and weight decay resolves them into the grokking transition. In the band setting those directions are orthogonal group characters carrying no target energy by construction. Weight decay removes them without surfacing a delayed correct solution, because no delayed correct solution is hiding there. The over-parameterization that grokking needs, structured features convert into overfitting. The sensitivity of grokking to experimental setting is well known, and what this adds is a cause for it.

The collapse curve of Figure~\ref{fig:main} (left) can be read as a restatement of double descent, with over-parameterization past the interpolation threshold degrading test error and the mechanism reducing to classical bias--variance. That signal-free directions can hurt generalization is classical, and we do not claim it. The specific claim here is dynamical. XVS establish that in the same regime an unstructured feature map produces a memorize-then-generalize \emph{delay} of order $1/\lambda$. A structured feature map, at fixed realizability and over-parameterization, does not shift that curve. It \emph{removes the delay}, leaving either gapless generalization or collapse and nothing in between (Appendix~\ref{app:transient}). A teacher-relevant direction with small positive $\mu$ is a slow clock and not a permanent obstruction, and positive weight decay attenuates such directions before they emerge. We therefore expect the delay to return as $\lambda \to 0$, on a timescale set by the smallest teacher-weighted eigenvalue and not by the weight decay, and preliminary observation at $\lambda = 0$ is consistent with a long memorize-then-generalize separation at the band radii that collapse here. Our claim concerns the positive-ridge regime, where the delay is absent and not merely long. The zero-ridge limit is treated in the full-length version. Double descent describes where the test-error curve sits. This concerns whether the delay appears at all, which makes the two compatible but distinct.

The static representability framework is inherited from Kam et al.'s holomorphic model~\citep{kam_holo}, and the symmetric-Laurent-band extension is a contribution of the present work. Solvable models of grokking have been studied in linear estimators~\citep{levi2024} and for modular addition in two-layer quadratic networks~\citep{mohamadi2024}, but these operate in the regime where memorization is always possible and the feature map is unstructured or learned. The present result is complementary to XVS~\citep{xvs2026}, delimiting the reach of their Gaussian-feature theorem by exhibiting a structured feature map on which its central phenomenon does not appear.

\section{Conclusion}

On a band-limited character feature map, a target the class provably contains is not learned by a memorize-then-generalize transition. Under positive weight decay it is learned at once when the active band is small and not at all when the active band is large, and the switch is governed by which modes are active, not by how many parameters there are.

The read-out is linear over a fixed feature map, so the two-layer jointly trained problem is untouched. The target carries a single mode, and a multi-mode target such as $2a-3b$ would activate a broader signal-relevant support, moving the threshold in a direction we have not mapped. Every run uses one weight decay, $\lambda = 10^{-2}$, leaving open the zero-ridge limit, where slow teacher-relevant directions stop being attenuated. The threshold, which Appendix~\ref{app:threshold} brackets between $2401$ and $3249$ active modes at this $(\lambda, n)$, is observed, not derived. Deriving it from $\lambda$, $n$, and band geometry is the next step, and Eq.~\eqref{eq:ridgelaw} says where to look, since a closed form needs a limiting description of the teacher-weighted spectral measure.

\bibliography{refs}

\appendix

\section{The transient at $M=24$ is onset of degradation, not grokking}
\label{app:transient}

\begin{figure}[h]
\centering
\includegraphics[width=0.62\textwidth]{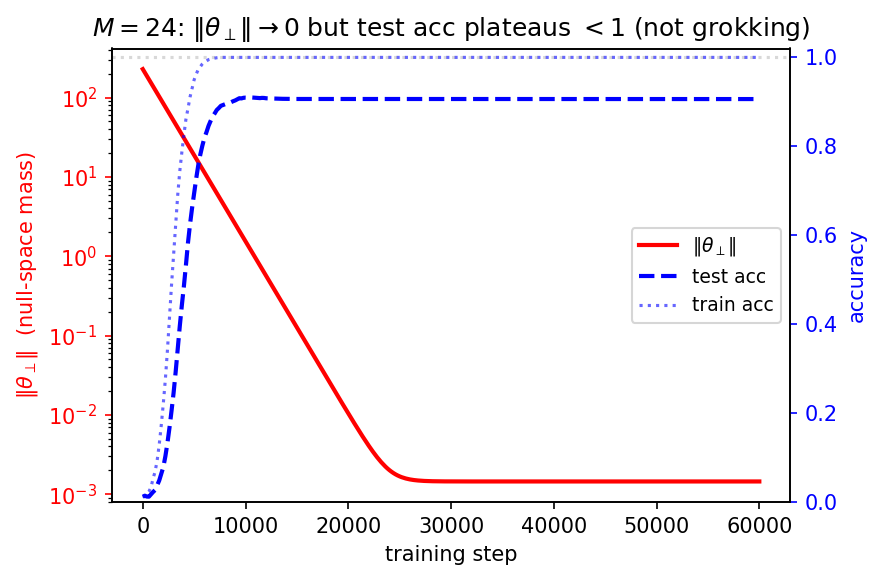}
\caption{Single-mode target $a-b$ at $M=24$ ($q/n=0.638$). The null-space mass $\|\theta_\perp\|$ (red, log scale) decays to numerical zero, but held-out accuracy (blue dashed) plateaus at $0.91$ and never reaches one, so the resolution of the null space is \emph{not} accompanied by a delayed rise in generalization. This distinguishes the transient from grokking, in which the correct solution surfaces as the null-space component decays.}
\label{fig:perp}
\end{figure}

The single point in Figure~\ref{fig:main} (left) that superficially resembles grokking is $M=24$ ($q/n=0.638$), where training accuracy still reaches $1.00$ while held-out accuracy has fallen to $0.91$. A train--test gap is the nominal signature of grokking, so the transient has to be separated from a memorize-then-generalize delay. This appendix does that in three steps, an operational criterion that already rules the point out, a rank argument showing that the mechanism XVS rely on cannot operate at this band radius, and the trajectory of Figure~\ref{fig:perp}, which confirms both.

\paragraph{The criterion is not met.} The grokking signature defined in Section~\ref{sec:setup} is a finite positive gap $t_2 - t_1$ at the $0.99$ threshold. At $M=24$ the peak held-out accuracy over the whole run is $0.912$, which never crosses the threshold, so $t_2$ does not exist and $t_2 - t_1$ is undefined. A delayed transition that never arrives is not a long delay. The same holds at every larger band radius, where peak accuracy is lower still. The nominal train--test gap at $M=24$ is therefore a gap in the sense of two numbers differing, not in the sense the grokking literature uses.

\paragraph{Two null spaces, only one of which exists here.} Write $\Phi_{\mathrm{tr}} = U S V^\top$ for the thin SVD and $P_\parallel = V_r V_r^\top$ for the projector onto the row space of rank $r$, so that the component invisible to the training data is $\theta_\perp = (I - P_\parallel)\theta$. Under XVS, grokking is driven by the slow decay of a \emph{signal-bearing} $\theta_\perp$, the delayed generalization occurring as weight decay resolves a correct solution hidden there. Whether such a solution can exist depends on where the null space comes from, and on this feature map there are two separate sources.

The first is the duplicated realification of Eq.~\eqref{eq:dim}. The implementation instantiates $2q$ channels spanning a space of dimension $q$, so $q$ coordinate directions are pure gauge and carry no function whatever the sample. The second is sampling, and it appears only once $q > n$, with dimension $q - n$.

At $M=24$ only the first source is present. Here $q = 2401 < n = 3763$. We verify numerically that $\Phi_{\mathrm{tr}}$ attains full column rank $2401$ on the deduplicated basis, with smallest singular value $2.08$ against a largest of $70.7$, so the rank is resolved with a wide margin and is not a threshold artefact. The complement of the row space inside the $4802$ implemented coordinates therefore has dimension $4802 - 2401 = 2401$, which is the gauge dimension. Every direction weight decay grinds down is a direction in which the predictor does not move. No delayed solution can be hiding in them, and the XVS mechanism has nothing to act on. The situation changes only above the threshold, where $M=32$ gives $q - n = 462$ genuinely unsampled functional directions, and those are the ones that produce the irreducible floor discussed in Section~\ref{sec:overfit}.

The decay of $\|\theta_\perp\|$ at $M=24$ is therefore a statement about coordinates and not about predictions, so it cannot be read as evidence of a resolving null space. The transient is likewise not a consequence of over-parameterization, since at $M=24$ the model is under-parameterized on the axis that matters.

\paragraph{The trajectory.} Figure~\ref{fig:perp} makes this concrete for the single-mode target $a-b$ at $M=24$ ($\lambda=10^{-2}$, $\nu=5$, learning rate $0.05$). The null-space mass $\|\theta_\perp\|$ decays monotonically from $228$ to $10^{-3}$ over roughly $25{,}000$ steps, while held-out accuracy reaches its ceiling of $0.91$ by step $\sim\!10{,}000$ and then does not move. Throughout the second half of training, while $\theta_\perp$ is still being ground to zero, held-out accuracy is flat. There is no late upswing coinciding with the resolution of the null space, and that upswing is the defining signature of grokking.

The reading is falsifiable in the obvious way. Had the curve risen late, in the window where $\|\theta_\perp\|$ crosses through the values that weight decay is removing, the XVS account would have been the right one even at this band radius. It does not, and the rank arithmetic above says why it could not have. We read the $M=24$ gap as the onset of degradation, with training accuracy saturating first, held-out accuracy already short of one, and both worsening as the band grows. The transient is not reported as grokking anywhere in the main text.

\paragraph{Scope.} The diagnostic is run at one band radius, one seed, and one weight decay, with the rank convention recorded in Appendix~\ref{app:repro}. The gauge argument is exact and needs neither of these choices. The trajectory is a single realization and is offered as confirmation, not as the load-bearing evidence.

\section{Closed-form representability in the character basis}
\label{app:closedform}

This appendix records why the spectral gap $\gap$ is cheaply computable, a property specific to the group-character feature map, and reports the numerical check summarized in Section~\ref{sec:setup}.

\begin{table}[h]
\centering
\begin{tabular}{lccccc}
\toprule
Target & $M$ & $q$ & $\gap$ & train acc & held-out acc \\
\midrule
$a+b$ & 1  & 9    & $7.0\times10^{-13}$ & 1.000 & 1.000 \\
$a+b$ & 24 & 2401 & $1.7\times10^{-17}$ & 1.000 & 1.000 \\
$a-b$ & 1  & 9    & $1.3\times10^{-12}$ & 1.000 & 1.000 \\
$a-b$ & 24 & 2401 & $1.8\times10^{-17}$ & 1.000 & 1.000 \\
$ab$  & 1  & 9    & $1.00$ & 0.013 & 0.012 \\
$ab$  & 3  & 49   & $1.00$ & 0.011 & 0.012 \\
$ab$  & 24 & 2401 & $2.54$ & 0.109 & 0.038 \\
\bottomrule
\end{tabular}
\caption{Closed-form ridge solution across targets and band radii. Representable targets have $\gap$ at numerical zero. The full-plane target $ab$ has $\gap = 1$ at every band, and the $\gap > 1$ entry is the finite-sample effect discussed at the end of this appendix. Dimensions are functional ($q$), per Eq.~\eqref{eq:dim}.}
\label{tab:rep}
\end{table}

Let $G = \Zp^2$ with characters $\chi_{u,v}(a,b) = \omega^{ua+vb}$, orthonormal under the normalized inner product $\langle f,g\rangle = |G|^{-1}\sum_{(a,b)} f(a,b)\overline{g(a,b)}$. The band feature map spans the subspace $\mathcal{F}_A = \operatorname{span}\{\chi_{u,v} : (u,v)\in A\}$ with $A = \{|u|,|v|\le M\}$. Because the characters are orthonormal, for any target $\omega^f$ the best band approximation is the orthogonal projection onto $\mathcal{F}_A$, and the residual energy is the out-of-band Fourier mass:
\begin{equation}
\gap \;=\; \big\| \omega^f - \operatorname{Proj}_{\mathcal{F}_A}\omega^f \big\|^2 \;=\; \sum_{(u,v)\notin A} \big|\widehat{\omega^f}(u,v)\big|^2 .
\label{eq:delta2}
\end{equation}
The quantity is static and training-free, and representability ($\gap = 0$) is decided by whether $\operatorname{supp}(\widehat{\omega^f}) \subseteq A$.

A point of bookkeeping applies to every row of Table~\ref{tab:rep} and not only to the anomalous one. Equation~\eqref{eq:delta2} is the residual of the exact orthogonal projection onto $\mathcal{F}_A$, computed against the full population of $p^2$ pairs. The tabulated numbers are instead the residual of the ridge solution fitted on a subsample of $n$ points at $\lambda = 10^{-6}$ and then evaluated on all $p^2$ pairs, normalized by the target energy. The two agree to numerical precision when the target is representable, because the projection is then exact and the fit recovers it. For an unrepresentable target the two separate, and the difference is the over-sample error discussed at the end of this appendix. We report the fitted quantity because it is what the experiments compute, and we give the population values below wherever they differ.

The accuracy columns are angular. A predicted phase is decoded to the nearest class, so $\gap$ and accuracy are different functionals of the same residual and need not move together in general. On the rows tabulated here they do. The residual is either at numerical zero, which gives exact recovery, or comparable to the total target energy, which gives chance.

\paragraph{Linear rules.} A linear-phase target $ma+nb$ is the single character $\chi_{m,n}$, so its Fourier support is the point $(m,n)$ and $\gap = 0$ iff $|m|,|n|\le M$. The signed rule $a-b = \chi_{1,-1}$ is representable at every $M\ge 1$, the extension that inverse phases (the two-channel Laurent construction) provide over a positive Fourier line, and Table~\ref{tab:rep} confirms it at $\gap\sim 10^{-12}$ already at $M=1$. The main text relies on one consequence. Any dynamical failure of $a-b$ at small $M$, and we do observe such failures under a nonlinear read-out, is an optimization artifact and not a representability barrier.

\paragraph{The multiplication rule.} For $f = ab$, the Fourier coefficient is
\begin{equation}
\widehat{\omega^{ab}}(u,v) = \frac{1}{p^2}\sum_{a,b} \omega^{ab - ua - vb} = \frac{\omega^{-uv}}{p},
\end{equation}
using $\sum_a \omega^{a(b-u)} = p\,[b\equiv u]$ and then summing over $b$. Every coefficient has modulus $1/p$ regardless of $(u,v)$, so the spectrum of $\omega^{ab}$ is flat over the whole plane $\Zp^2$. The derivation uses nothing beyond character orthogonality, and the flatness is the discrete counterpart of a chirp being spread across all frequencies.

First, the in-band energy fraction is $|A|/p^2 = q/p^2$, so the population residual is
\begin{equation}
\gap_{\mathrm{pop}}(ab) = 1 - \frac{q}{p^2},
\label{eq:abpop}
\end{equation}
which equals $0.9990$, $0.9948$ and $0.7448$ at $M = 1, 3, 24$. Second, $\gap_{\mathrm{pop}} = 0$ requires $q = p^2$, that is $M = (p-1)/2 = 48$ and a band equal to the entire character basis. There is no partial remedy. A model that represents $ab$ on $\Zp^2$ must be able to represent every function on $\Zp^2$, which is why $ab$ is the irreducible obstruction of the parent holomorphic model and not a target that a wider band would eventually reach.

\paragraph{Why $\gap$ can exceed one.} Comparing the fitted values against Eq.~\eqref{eq:abpop} shows that the excess is present on every unrepresentable row and merely becomes conspicuous on one of them.

\begin{center}
\begin{tabular}{cccc}
\toprule
$M$ & $q$ & fitted $\gap$ & population $\gap_{\mathrm{pop}}$ \\
\midrule
1  & 9    & $1.0004$ & $0.9990$ \\
3  & 49   & $1.0038$ & $0.9948$ \\
24 & 2401 & $2.5365$ & $0.7448$ \\
\bottomrule
\end{tabular}
\end{center}

The fitted residual exceeds the population residual at all three band radii, by $10^{-3}$ at $M=1$ and by a factor of $3.4$ at $M=24$, and the gap widens with $q$. The reason is that the fitted coefficients minimize error on a subsample of $n$ points and not on the population, so they differ from the orthogonal projection, and the off-sample error they incur adds to rather than subtracts from the residual of Eq.~\eqref{eq:delta2}. The measured quantity is an over-sample error and is not bounded by the target energy, which is why a number above one is not a contradiction.

This is not an interpolation effect. At $M=24$ the functional dimension is $q = 2401 < n = 3763$, so the system is overdetermined and the training set cannot be interpolated. What grows with $q$ is the number of in-band directions available to absorb subsample-specific structure, and at $\lambda = 10^{-6}$ almost nothing restrains them. The same mechanism is what Section~\ref{sec:overfit} exhibits dynamically for a \emph{representable} target, where it costs generalization instead of showing up as an inflated residual.

\section{A heuristic for the collapse threshold}
\label{app:threshold}

We give an order-of-magnitude account of where the collapse threshold should sit, and then bracket it against the data. This is a heuristic, not a theorem, and a rigorous treatment is deferred.

Let $m_{\mathrm{act}}$ be the number of active modes, of which one is signal-relevant (the target mode) and $m_{\mathrm{act}} - 1$ are signal-free. On the training subsample of size $n$, the ridge solution fits both the target mode and the empirical projections of the signal-free modes, and the latter contribute out-of-sample error controlled by their finite-sample aliasing. For orthonormal characters $\chi_j$ and $\chi_t$ with $j \ne t$, the empirical inner product $n^{-1}\sum_{x \in S}\chi_j(x)\overline{\chi_t(x)}$ has mean zero, since each summand has mean zero over the population. Its second moment is $n^{-2}$ times the variance of the sum. The summands are uncorrelated and of unit modulus, so that variance is $n$, giving
\begin{equation}
\mathbb{E}\big|\langle \chi_j, \chi_t\rangle_S\big|^2 = \frac{1}{n}\,\Big(1 - \frac{n-1}{N-1}\Big) \le \frac{1}{n},
\label{eq:alias}
\end{equation}
where the bracket is the finite-population correction for sampling without replacement and equals $0.60$ at our $n/N$. Summing over the $m_{\mathrm{act}}-1$ signal-free modes gives an aliasing energy of order $(m_{\mathrm{act}}-1)/n$. Weight decay $\lambda$ shrinks each spurious coefficient, suppressing this energy by a factor that saturates once $m_{\mathrm{act}}/n$ is order one. The transition to collapse is therefore expected near
\begin{equation}
m_{\mathrm{act}}^{\star} \;\sim\; c(\lambda)\, n,
\end{equation}
with $c(\lambda)$ an $O(1)$ factor increasing with the regularization strength.

\paragraph{Bracketing the constant.} The masking sweep alone leaves a wide window, since it samples only $1089$ (perfect) and $4225$ (collapsed). The capacity sweep supplies intermediate points at the same $\lambda$ and the same $n$, and the two can be placed on a single axis of active modes. Doing so is not neutral. It presumes the claim of Section~\ref{sec:overfit}, that only the active support matters and the nominal dimension does not, so a masked band with $m_{\mathrm{act}}$ active modes should behave like an unmasked band with $q = m_{\mathrm{act}}$, and where the two sweeps nearly meet they do.

\begin{center}
\begin{tabular}{rcl}
\toprule
$m_{\mathrm{act}}$ & peak held-out acc & source \\
\midrule
$1089$ & $1.000$ & masked, $k=16$ \\
$1369$ & $0.999$ & unmasked, $M=18$ \\
$1849$ & $0.992$ & unmasked, $M=21$ \\
$2401$ & $0.912$ & unmasked, $M=24$ \\
$3249$ & $0.506$ & unmasked, $M=28$ \\
$4225$ & $0.185$ & both, $M=32$ \\
\bottomrule
\end{tabular}
\end{center}

The two independent routes to $4225$ agree, and $1089$ masked sits beside $1369$ unmasked with a difference of $0.001$. The combined curve narrows the window considerably. Measurable degradation begins near $m_{\mathrm{act}} = 2401$, which is $0.638\,n$, and half-height is reached near $3249$, which is $0.863\,n$. These are the same two ratios that appear as $q/n$ in Table~\ref{tab:collapse}, since the axis is the same. Reading the half-height point as the threshold gives $c(10^{-2}) \approx 0.86$, which is the order-unity constant the argument predicts without fixing.

\paragraph{What this does not establish.} The scaling $m^{\star}_{\mathrm{act}} \propto n$ is untested, since every run here uses one $n$. The dependence of $c$ on $\lambda$ is likewise untested, since every run uses $\lambda = 10^{-2}$, so the claim that $c$ increases with regularization is an expectation from Eq.~\eqref{eq:alias} and not a measurement. The same character second-moment computation controls both the aliasing energy and the conditioning of the empirical Gram matrix, and we expect a full-length analysis, via matrix concentration bounds on the empirical Gram~\citep{tropp2015}, to yield $c(\lambda)$ in closed form.

Finally, the count is a proxy. The quantity appearing above is $m_{\mathrm{act}}$, the number of activated signal-free modes, and not the nominal dimension, which is the distinction the masking experiment isolates. By Eq.~\eqref{eq:ridgelaw}, however, the count acts on the error only through the teacher-weighted spectrum, and Eq.~\eqref{eq:alias} averages over modes where the exact object does not. That is why we present this as a scaling and not as a law, and why the bracket above is a description of one $(\lambda, n, \text{band})$ configuration rather than a threshold formula.

\section{Experimental detail and measured values}
\label{app:repro}

Tables~\ref{tab:collapse} and~\ref{tab:mask} are the numbers plotted in Figure~\ref{fig:main}, with seed-to-seed standard deviations, and the protocol that produced them is recorded below.

\begin{table}[h]
\centering
\begin{tabular}{cccc}
\toprule
$M$ & $q$ & $q/n$ & peak held-out acc (mean $\pm$ std) \\
\midrule
18 & 1369 & 0.364 & $0.999 \pm 0.001$ \\
21 & 1849 & 0.491 & $0.992 \pm 0.002$ \\
24 & 2401 & 0.638 & $0.912 \pm 0.006$ \\
28 & 3249 & 0.863 & $0.506 \pm 0.006$ \\
32 & 4225 & 1.123 & $0.185 \pm 0.006$ \\
40 & 6561 & 1.744 & $0.066 \pm 0.004$ \\
\bottomrule
\end{tabular}
\caption{Peak held-out accuracy versus capacity for the representable target $a-b$, with $n=3763$. Accuracy falls monotonically toward chance as $q/n$ grows. The first two collapse points, $M=24$ and $M=28$, lie \emph{below} the interpolation threshold $q/n=1$.}
\label{tab:collapse}
\end{table}

\begin{table}[h]
\centering
\begin{tabular}{ccc}
\toprule
keep radius $k$ & active modes & peak held-out acc (mean $\pm$ std) \\
\midrule
1  & 9    & $1.000 \pm 0.000$ \\
2  & 25   & $1.000 \pm 0.000$ \\
4  & 81   & $1.000 \pm 0.000$ \\
8  & 289  & $1.000 \pm 0.000$ \\
16 & 1089 & $1.000 \pm 0.000$ \\
32 & 4225 & $0.185 \pm 0.006$ \\
\bottomrule
\end{tabular}
\caption{Masking at fixed nominal dimension, $M=32$, with masked channels zeroed and not removed. Active modes are counted functionally as $(2k+1)^2$. Restricting the active support to any radius up to $1089$ modes, while keeping the target mode, gives perfect and zero-variance generalization. Only the full $4225$-mode band collapses.}
\label{tab:mask}
\end{table}

\paragraph{Setup.} All experiments use $p=97$. Features are the real ($\cos$/$\sin$) channels of the band characters. The implementation instantiates all $2(2M+1)^2$ channels over the full symmetric band. By Eq.~\eqref{eq:dim} these span a space of functional dimension $q = (2M+1)^2$, and every capacity figure quoted in this paper is the functional value $q$. The masked variant zeros all channels outside keep-radius $k$, symmetrically in $\pm(r,s)$, while preserving the number of instantiated columns. A single train/test split at fraction $0.4$ is drawn per seed. The seed is set once, a uniform permutation of the $p^2 = 9409$ cells is drawn, and its first $n = \lfloor 0.4\,p^2\rfloor = 3763$ entries form the training set. The seed therefore fixes both the split and the initialization, and every run is reproducible from it alone. Held-out accuracy is measured on the remaining $5646$ cells.

\paragraph{Null-space diagnostic.} The trajectory of Appendix~\ref{app:transient} is a separate run at $M=24$ with the same hyperparameters and seed $0$. The thin SVD of $\Phi_{\mathrm{tr}}$ is taken once before training, and $\|\theta_\perp\|$ is recorded on the same $100$-step grid as the accuracies. The rank $r$ is fixed by a relative threshold of $10^{-8}$ on the leading singular value.

\paragraph{Penalty normalization.} Weight decay is applied uniformly to all $2(2M+1)^2$ instantiated coordinates. Because each function is represented by a conjugate pair of coordinates that gradient descent loads symmetrically, a coefficient of size $c$ in the deduplicated orthonormal basis is penalized as $2(c/2)^2 = c^2/2$ in the implemented one. The quoted $\lambda = 10^{-2}$ is therefore the value in the implemented parameterization and corresponds to $5\times10^{-3}$ in the deduplicated basis. The two parameterizations span the same predictors, but numerical values of $\lambda$ are not comparable across them, and we carry the implemented value throughout. Closed-form results solve $(\Phi_{\mathrm{tr}}^\top\Phi_{\mathrm{tr}} + \lambda I)\theta = \Phi_{\mathrm{tr}}^\top Y$ with $\lambda = 10^{-6}$ and report the population residual over all $p^2$ points. Dynamical results use full-batch gradient descent on the mean-squared error. Weight decay is supplied by the optimizer and not by an explicit penalty term (SGD, learning rate $0.05$, weight decay $\lambda = 10^{-2}$, no momentum). The read-out is a single matrix $\theta \in \mathbb{R}^{2(2M+1)^2 \times 2}$ with independent $\mathcal{N}(0,\nu^2)$ entries at $\nu = 5$ and no bias term. The target phase is regressed in $(\cos,\sin)$ coordinates, and a prediction counts as correct when its angle lies within $\pi/p$ of the target angle, the half-width of the decision wedge for $p$ classes. Figure~\ref{fig:main} uses three seeds $\{0,1,2\}$. Peak held-out accuracy is the best value attained at any evaluation point over the whole run, sampled every $100$ steps. It is an upper envelope and not a terminal value, so it lies slightly above the ridge fixed point of Eq.~\eqref{eq:ridgelaw}. The two panels use different budgets: $60{,}000$ steps for the capacity sweep and $30{,}000$ for the masking sweep. The budgets are consistent where they overlap: the $k=32$ masking configuration is the unmasked $M=32$ band and reaches $0.185$ under both, indicating that $30{,}000$ steps already suffice for convergence in that regime. Every number reported in this paper follows from these settings and the listed seeds, with no per-configuration tuning. Code and seeds are available from the authors upon request.